\documentclass[letterpaper]{article} 
\usepackage{aaai2026}  
\usepackage{times}  
\usepackage{helvet}  
\usepackage{courier}  
\usepackage[hyphens]{url}  
\usepackage{graphicx} 
\usepackage{natbib}  
\usepackage{caption} 
\usepackage{newfloat}
\usepackage{listings}
\DeclareCaptionStyle{ruled}{labelfont=normalfont,labelsep=colon,strut=off} 
\newcommand{\acro}{\textbf{\texttt{TRACE}}\xspace}
\newcommand{\internvl}{\textsc{InternVL-2.5-8B}\xspace}
\newcommand{\gemini}{\textsc{Gemini-2.5-flash}\xspace}
\newcommand{\gpt}{\textsc{GPT-4o-mini}\xspace}
\newcommand{\cliplarge}{\textsc{CLIP-ViT-L/14}\xspace}

\newcommand{\clipxlmr}{\textsc{CLIP-XLM-R-ViT-H/14}\xspace}

\newcommand{\siglip}{\textsc{SigLIP2-L/16-384}}

\usepackage{booktabs}
\usepackage{multirow}
\usepackage{makecell}
\usepackage{array}
\usepackage{xspace}

\usepackage{longtable}            
\usepackage{subcaption}
\usepackage{tikzsymbols}
\usepackage{bbding}
\usepackage{listings}
\usepackage{color}
\usepackage{resizegather}

\usepackage[hashEnumerators,smartEllipses]{markdown}
\usepackage[export]{adjustbox}

\title{\acro: Textual Relevance Augmentation and Contextual Encoding for Multimodal Hate Detection}
\author {
    Girish A. Koushik\textsuperscript{\rm 1}\thanks{Corresponding author},
    Helen Treharne\textsuperscript{\rm 2},
    Aditya Joshi\textsuperscript{\rm 3},
    Diptesh Kanojia\textsuperscript{\rm 1}
}
\affiliations {
    \textsuperscript{\rm 1}Nature-Inspired Computing \& Engineering, University of Surrey, Guildford, United Kingdom\\
    \textsuperscript{\rm 2}Surrey Centre for Cyber Security, University of Surrey, Guildford, United Kingdom\\
    \textsuperscript{\rm 3}University of New South Wales, Sydney, Australia\\
    g.koushik@surrey.ac.uk, h.treharne@surrey.ac.uk, aditya.joshi@unsw.edu.au, d.kanojia@surrey.ac.uk
}

\begin{document}

\maketitle

\begin{abstract}
Social media memes are a challenging domain for hate detection because they intertwine visual and textual cues into culturally nuanced messages. To tackle these challenges, we introduce \acro, a hierarchical multimodal framework that leverages visually grounded context augmentation, along with a novel caption-scoring network to emphasize hate-relevant content, and parameter-efficient fine-tuning of CLIP’s text encoder. Our experiments demonstrate that selectively fine-tuning deeper text encoder layers significantly enhances performance compared to simpler projection-layer fine-tuning methods. Specifically, our framework achieves state-of-the-art accuracy ($0.807$) and F1-score ($0.806$) on the widely-used Hateful Memes dataset, matching the performance of considerably larger models while maintaining efficiency. Moreover, it achieves superior generalization on the MultiOFF offensive meme dataset (F1-score $0.673$), highlighting robustness across meme categories. Additional analyses confirm that robust visual grounding and nuanced text representations significantly reduce errors caused by benign confounders. We publicly release our code to facilitate future research.



\textbf{Disclaimer}: This paper includes references to potentially disturbing, hateful, or offensive content due to the nature of the task.
\end{abstract}

\begin{links}
    \link{Code}{https://github.com/gak97/TRACE}
\end{links}

\section{Introduction} \label{sec:introduction}
Social media platforms provide the environment by which multimedia content, such as internet memes, can proliferate, evolve swiftly, making them difficult to moderate~\cite{young2022much}. Internet memes represent benign, humorous, or satirical images combined with a caption in the form of overlaid text~\cite{denisova2019internet}. In hateful memes, the images are repurposed to propagate hateful messages, reinforce stereotypes, and target individuals and communities. Detecting hateful memes requires not only an understanding of the visual content but also the underlying linguistic and cultural context that distinguishes hateful rhetoric from benign humour. 

Understanding hateful memes poses two major challenges for current vision–language models. Firstly, memes are not simply a juxtaposition of an image with overlaid text; instead, they convey meaning through a nuanced fusion in which the image can provide context, irony, or subtext to the accompanying words in the text. Hence, models need to capture subtle semantic and cultural cues within mathematical representations. This remains difficult for the current state-of-the-art architectures \citep{kiela2020hateful}. 

\begin{figure}[!t] 
    \centering
    \includegraphics[width=7.5cm]{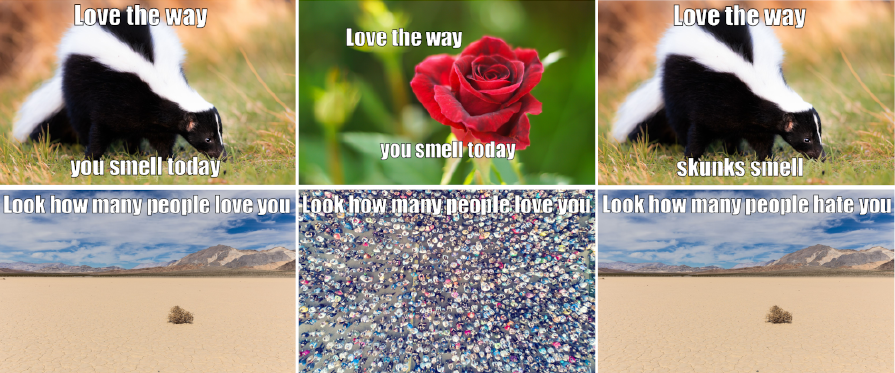}
    \caption{Illustration of benign confounders (absent from the dataset) as noted by~\citet{kiela2020hateful}. (left) meme, (centre) image confounder, (right) text confounder.}
    \label{fig:benco_illustration}
\end{figure}

Secondly, models often misclassify content because of the impact that benign elements have on representations. For example, in Figure \ref{fig:benco_illustration}  (top row, left-most image), the text along with the image yields a hateful meme. However, the benign image along with the same text (top row, middle image) yields a non-hateful meme. Such combinations are known as ``benign confounders'' \citep{kiela2020hateful}. While this is an example of a textual confounder, similar image confounders also exist. These confounders prevent models from exploiting a combination of unimodal representations, \textit{i.e.}, neither the visual modality nor the textual modality alone reliably indicates the presence of hate \citep{aggarwal2024text,koushik2025robustframeworkmultimodalhate}. As a result, there is a need for advanced multimodal strategies for the detection of hateful memes.

\begin{figure}[!t]
    \centering
    \includegraphics[width=\columnwidth]{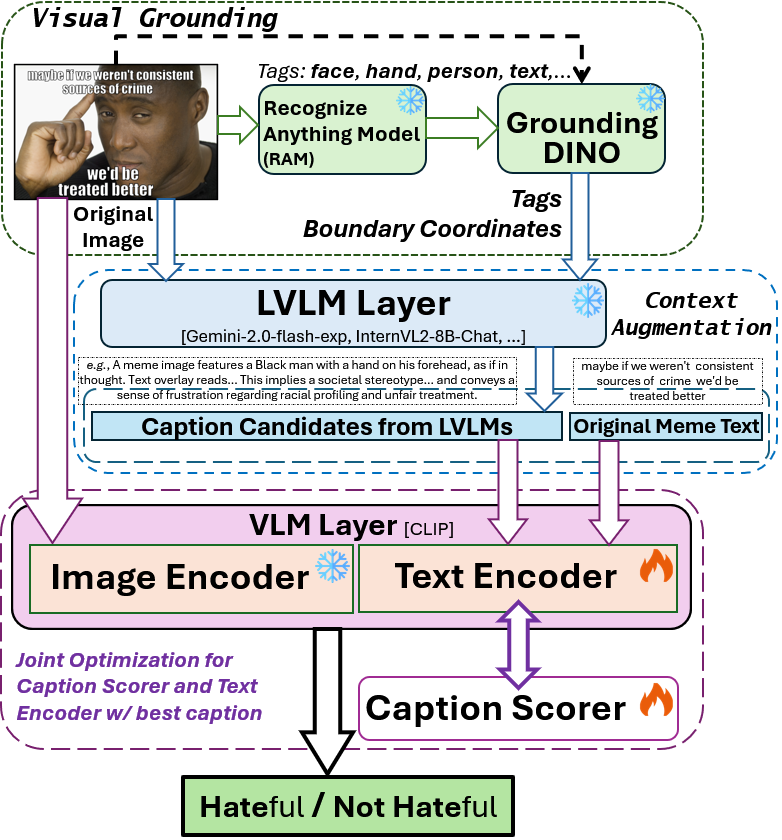}
    \caption{Proposed hierarchical architecture of the \acro framework for Hateful Meme detection. The process includes: (1) Visual Grounding using RAM and GroundingDINO; (2) Context Augmentation where LVLMs generate captions incorporating grounded visual details and original text; (3) A VLM layer (such as CLIP) encoding the image and augmented text; (4) A novel Caption Scorer selecting the most relevant caption; and (5) Joint fine-tuning the text encoder and caption scorer for final classification. Frozen components (\Snowflake) remain untrained.}
    \label{fig:main_arch}
\end{figure}

Existing approaches to hateful meme detection span a diverse range of multimodal techniques, each attempting to reconcile the conflicting and often subtle signals present in meme images and their overlaid text \cite{hu2023visual,mei2023improving,kumar2022hate}. Many of these approaches achieve good results, but they focus on either capturing the complex semantic interplay inherent in memes or mechanisms for distinguishing benign confounders. To understand memes, we believe a framework needs to encapsulate a holistic understanding of the image and text combination along with cultural context, which may be missing from the original meme text. To this end, we propose a hierarchical interpretable framework, \acro, that performs \textit{multimodal context augmentation}, while enriching input information related to the meme image prior to augmentation. The \textbf{main contributions} of this paper are:

\begin{itemize}
    \item A hierarchical, interpretable framework, referred to as \acro, for hate detection (as shown in Figure~\ref{fig:main_arch}) that includes visual grounding to reduce the complexity of meme images when performing multimodal context augmentation.
    \item 
    Within \acro, we propose an efficient \textit{n}-layer fine-tuning approach, and novel joint optimization of vision-language models using a caption-scoring neural network. The caption scorer takes candidate captions as input and selects the best image-caption pair during the fine-tuning process.
    \item A comprehensive evaluation of various model configurations with both quantitative and qualitative error analysis. Notably, \acro attains high accuracy ($0.807$) and an F1 score ($0.806$) on the Hateful Memes dataset, at par with the existing SoTA framework while being more efficient.
\end{itemize}

\section{Related Work} \label{sec:background}
\paragraph{Multimodal Hate Detection} Recent advances in multimodal hate detection have shown promising developments in analysing hateful memes, though significant challenges persist. \citet{gomez2020exploring} established early foundations with the MMHS150K dataset, implementing a dual-stream architecture combining Inceptionv3 \citep{szegedy2016rethinking} for visual features and LSTM \citep{hochreiter1997long} for textual features achieving a modest accuracy of $68.2\%$. The field then shifted toward contrastive learning approaches: HateSieve \citep{su2024hatesieve} combined a large vision–language model (LVLM) for caption generation with SDXL \citep{meng2021sdedit} for image synthesis, achieving $73.45\%$ accuracy on the FHM dataset \citep{kiela2020hateful}; Hate-CLIPper \citep{kumar2022hate} fine-tuned Contrastive Language-Image Pre-training (CLIP)’s projection layers for cross-modal interaction to reach an AUROC score of $0.858$ (Acc $0.756$), which was further improved by Retrieval-guided Contrastive Learning (RGCL) \citep{mei2023improving} to AUROC $0.870$ (Acc $0.788$) through a runtime retrieval database. While these methods outperform earlier models, they still struggle with benign confounders and complex semantic relationships between modalities, as cosine-based similarity can yield inconsistent performance \citep{steck2024cosine}. More compute-intensive approaches like PALI-X-VPD \citep{hu2023visual} achieve state-of-the-art AUROC $0.892$ (Acc $0.808$) via a $55B$‑parameter language model with chain-of-thought prompting, raising concerns about real-world deployability. Alternative lightweight architectures such as fine-tuned Flamingo \citep{alayrac2022flamingo} and ISSUES \citep{burbi2023mapping} have reported competitive AUROC scores of $0.866$ and $0.855$, respectively. Domain-specific datasets have also emerged: MultiOFF \citep{suryawanshi2020multimodal} for the 2016 US elections (F1 $0.54$) and CrisisHateMM \citep{bhandari2023crisishatemm} for Russia–Ukraine conflict memes (F1 $0.786$).


Building on these foundations, recent work has explored richer semantic and knowledge-guided representations. \cite{zhong2024multimodal} introduces a fairness-aware vision–language framework that generates human-centric explanations for multimodal memes across domains, focusing on interpretability rather than classification accuracy. \cite{grasso2024kermit} presents KERMIT, a memory-augmented model integrating ConceptNet knowledge for harmful meme detection, reporting $85.3\%$ AUROC on the FHM dataset. \cite{lin2024towards} harnesses multimodal debates between large language models to generate conflicting rationales, then fine-tunes a lightweight judge model for harmfulness inference, improving explainability and detection robustness. These approaches suggest that integrating external knowledge, richer semantic signals, and fairness considerations can complement contrastive and large‑scale models, balancing detection accuracy with interpretability and deployment feasibility.

\paragraph{Caption Generation} The advent of vision‐language models (VLMs) has significantly transformed the task of image caption generation and caption expansion, enabling the production of descriptions that are both detailed and contextually enriched. Early approaches relied on encoder-decoder architectures, for example, models such as ``Show and Tell'' \citep{vinyals2015show} and ``Show, Attend and Tell'' \citep{xu2015show} used convolutional neural networks to extract visual features and recurrent neural networks to generate captions, but these methods were often limited to relatively brief descriptions. More recent work has leveraged transformer‐based architectures and large pre‐trained language models to achieve greater semantic fidelity; for instance, ClipCap \citep{mokady2021clipcap} integrates CLIP embeddings with a language model to generate captions that are more closely aligned with visual content. In addition, there have been LVLMs such as BLIP-2 \citep{li2023blip} and InternVL \citep{chen2024internvl} that have achieved state-of-the-art (SoTA) scores on caption generation tasks. For our work, we leverage the \internvl model for caption generation, as the meme text does not describe the image but rather forms the meaning in conjunction with the image.

\paragraph{Visual Grounding} Although LVLMs have a good understanding of images, they still hallucinate on images such as memes, which subtly convey their meaning. Therefore, providing additional context in the form of visual grounding helps the model ``see'' better \citep{bai2024hallucination}. Recent progress in VLMs has developed multiple methods that integrate visual grounding, improving the alignment between text inputs and their corresponding visual areas. This enhancement benefits applications such as visual question answering (VQA), image retrieval, and object localization. For instance, one line of work has focused on rephrasing and augmenting input queries based on salient visual content \citep{prasad2023rephrase}, which in turn improves zero‐shot performance on VQA tasks. In another approach, researchers have explored learning visual grounding from generative VLMs by eliciting object‐level descriptions through carefully designed prompts, effectively using pre-trained image-text alignment without relying on extensive manual annotations \citep{wang2024learning}. Several other works use visual grounding for object localization \citep{yang2023improving}, image compression \citep{liu2024tell}, and reducing hallucinations \citep{yan2024vigor} in VLMs. To reduce hallucination and improve caption quality, we leverage visual grounding in the form of object tagging and detection.

\section{Methodology} \label{sec:methodology}

\begin{figure}[!t]
    \centering
    \includegraphics[width=0.95\linewidth]{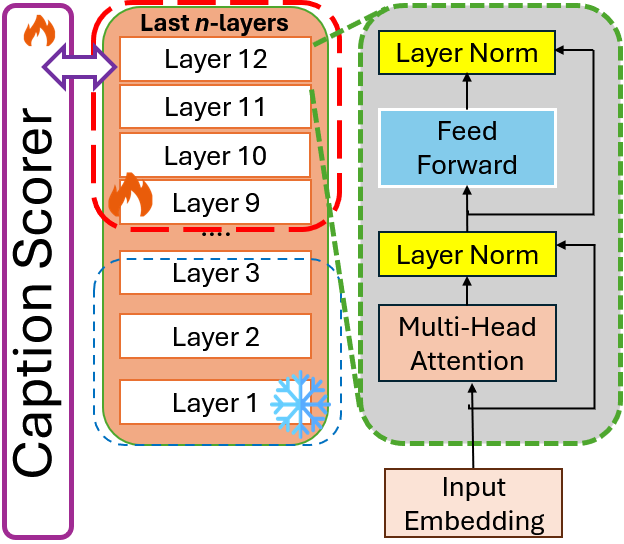}
    \caption{Parameter-Efficient Fine-tuning of the CLIP text encoder within \acro. This diagram illustrates the strategy where only the last $n$-layers (indicated by \Fire symbol) of the text encoder (comprising Layer Norm, Multi-Head Attention, and Feed Forward blocks) are fine-tuned, alongside the Caption Scorer. This minimizes trainable parameters to prevent overfitting while adapting the model for hate detection.}
    \label{fig:peft_arch}
\end{figure}

Our \acro framework consists of three blocks as shown in Figure~\ref{fig:main_arch}. We describe the first two blocks in $\S$~\ref{subsec:vgandca}, and the third block in $\S$~\ref{subsec:fine-tuning}. Figure~\ref{fig:peft_arch} provides a more detailed view of the third block. In both figures, the {\Snowflake} symbol indicates that framework components in \acro are frozen and thus remain untrained on our data. The {\Fire} symbol highlights that \acro proposes training for these components.

\subsection{Visually Grounded Context Augmentation} \label{subsec:vgandca}

To create visual grounding from the original image, we utilise the Recognize Anything Model (RAM) \citep{zhang2024recognize} for tagging, and GroundingDINO \citep{liu2023grounding} for detecting objects along with their bounding boxes. Although zero-shot object detection methods, \textit{e.g.}, YOLOv9 \citep{wang2024yolov9}, can produce bounding boxes for generic images, they fall short in capturing the context of hateful memes. Hence, we employ RAM for tag generation paired with GroundingDINO for open-vocabulary object detection. The next step is to augment the context from the image into the text. To do so, the tags from RAM and bounding box coordinates from GroundingDINO are then fed into the LVLM layer to generate captions. These captions highlight interactions between visual and textual elements, with an emphasis on cultural references essential for meme comprehension, as shown in the supplementary file. \acro supports the generation of captions using several different VLMs. All such captions and the original meme text are then encoded for the caption scorer, as described below.


\begin{table}[!ht]
    \centering
    \small
    \setlength{\tabcolsep}{4pt}
    \renewcommand{\arraystretch}{1.3}
    \begin{tabular}{>{\footnotesize\centering\arraybackslash}p{2.2cm}|*{4}{c}}
        \toprule
        \textbf{Model} & 
        \makecell{\textbf{\# of}\\\textbf{Params}} & 
        \makecell{\textbf{\# of}\\\textbf{Text Enc.}\\\textbf{Layers}} & 
        \makecell{\textbf{Text}\\\textbf{Enc.}} & 
        \makecell{\textbf{\#}\\\textbf{Last-4}} \\
        \midrule
        CLIP-ViT-L/14 & 428M & 12 & 123M & 41M \\[2pt]
        \siglip & 882M & 24 & 413M & 68M \\ 
        \makecell{CLIP-XLM-R-\\ViT-H-14} & 1.1B & 24 & 540M & 90M \\
        \bottomrule
    \end{tabular}
    \caption{Number (\#) of parameters for CLIP and SigLIP models. `Text Enc.' reports parameters from the text encoder, and `Last-4' reports parameters in the last four layers of the text encoder.}
    \label{tab:model-params}
\end{table}

\subsection{Caption-aware Text Encoder Fine-tuning} \label{subsec:fine-tuning}
The last block of \acro is a combination of a vision-language model and a novel neural network. \acro builds upon CLIP, a foundational vision-language encoder model which was trained using a contrastive learning approach~\citep{radford2021learning}. CLIP is jointly trained using image-text pairs, and our task requires an understanding of the nuanced interplay between visual content and text for detecting hateful memes, making it well-suited as a component. 

Since our training dataset contains limited samples (approx. $8.5k$), fine-tuning all layers in the text encoder (as shown in Table \ref{tab:model-params}) is likely to result in over-fitting on the training set, worsening performance on the validation and the test set. Therefore, for such a low-resource scenario, \acro adopts parameter-efficient fine-tuning \citep{houlsby2019parameter}, and proposes only tuning the last $n$-layers of the text encoder, as shown in Figure \ref{fig:peft_arch}. Additionally, our experiments show the viability of only tuning the text encoder, leaving the image encoder untrained. 

\acro feeds the candidate captions to the text encoder for obtaining representations, and further proposes the use of our novel caption scorer, \textit{i.e.}, a feed-forward neural network with $3$ hidden layers detailed as follows. 

The \textbf{caption scorer} $\mathcal{S}$ processes textual features \(h_i \in \mathbb{R}^d\) (where $d$ is the output dimension of the CLIP text encoder) to produce relevance scores $S_i$.
\[
\begin{aligned}
S_i = f_{\theta}(\mathbf{h}_i) &= \mathbf{W}_5 \bigg( \phi_4 \bigg( \mathbf{W}_4 \Big( \phi_3 \Big( \mathbf{W}_3 \Big( \phi_2 \big( \\
&\qquad \mathbf{W}_2 \big( \phi_1 (\mathbf{W}_1 \mathbf{h}_i) \big) \big) \Big) \Big) \Big) \bigg) \bigg)
\end{aligned}
\]
where:
\begin{gather*}
\phi(x) = \text{GELU}(\text{LayerNorm}(x)) \odot \text{Dropout}(p)\\
\mathbf{W}_1 \in \mathbb{R}^{d \times 1024}, \quad \mathbf{W}_2, \mathbf{W}_3 \in \mathbb{R}^{1024 \times 1024}, \mathbf{W}_4 \in \mathbb{R}^{1024 \times 512}, \quad \mathbf{W}_5 \in \mathbb{R}^{512 \times 1}
\end{gather*}

Weight normalisation \( \mathbf{W} = \frac{g}{\|\mathbf{v}\|} \mathbf{v} \) is applied to the weight matrices $\mathbf{W}_2$, $\mathbf{W}_3$, and $\mathbf{W}_4$. $\mathbf{W}_1$ is the input layer and contains only input features, while $\mathbf{W}_5$, on the other hand, is the output layer where weight normalization can be applied. However, in our case, we observed better training stability and performance without the normalization. Hence, we apply normalization only to $\mathbf{W}_2$-$\mathbf{W}_4$ layers. Bias terms are omitted for brevity, but applied in each linear layer.

Second, \textbf{Gumbel-Softmax} enables differentiable caption selection:
\[p_i = \frac{exp((s_i+g_i)/\tau}{\Sigma^n_{j=1}exp((s_j+g_j)/\tau}\]

where \(g_i \sim -log(-log(\mathcal{U}(0,1)))\) is Gumbel noise, $\tau$ is the temperature, and n is the number of caption candidates. The selected caption and image features are projected onto a larger projection space and then undergo bidirectional cross-attention fusion before classification. Given image features \(\mathbf{I}\) and text features \(\mathbf{T}\), the cross-attention mechanism computes:

\[\mathbf{I}_p = \text{Projection}(\mathbf{I}) \in \mathbb{R}^{B \times D}\] 
\[\mathbf{T}_p = \text{Projection}(\mathbf{T}) \in \mathbb{R}^{B \times D}\]
where $B$ is batch size and $D$ is projection dimension ($1024$).

\[\text{CrossAttn}(\mathbf{Q}, \mathbf{K}, \mathbf{V}) = \text{softmax}\left(\frac{\mathbf{Q}\mathbf{K}^T}{\sqrt{d_K}}\right)\mathbf{V}\]

For image-to-text attention:
\[\mathbf{I}_{\text{enhanced}} = \mathbf{I_p} + \text{CrossAttn}(\mathbf{I_p}, \mathbf{T_p}, \mathbf{T_p})\]
where $\mathbf{Q} = \mathbf{I}_p$ (Query is image), $\mathbf{K} = \mathbf{T}_p$ (Key is text), and $\mathbf{V} = \mathbf{T}_p$ (Value is text).

For text-to-image attention:
\[\mathbf{T}_{\text{enhanced}} = \mathbf{T_p} + \text{CrossAttn}(\mathbf{T_p}, \mathbf{I_p}, \mathbf{I_p})\]
where $\mathbf{Q} = \mathbf{T}_p$ (Query is text), $\mathbf{K} = \mathbf{I}_p$ (Key is image), and $\mathbf{V} = \mathbf{I}_p$ (Value is image).

The enhanced features are concatenated:
\[\mathbf{F}_{\text{combined}} = [\mathbf{I}_{\text{enhanced}} ; \mathbf{T}_{\text{enhanced}}] \in \mathbb{R}^{B \times 2D}\]

This bidirectional attention allows each modality to be enhanced by information from the other before classification.

Finally, a \textbf{hate relevance loss} \( \mathcal{L}_{\text{rel}} \) aligns caption selection with labels:

\begin{gather*}
\mathcal{L}_{\text{rel}} = - \left[ y \log \left( \sum p_i \right) + (1 - y) \log \left( 1 - \sum p_i \right) \right]
\end{gather*}

The classification loss ($\mathcal{L}_{\text{cls}}$) uses combined features of both the image and the selected caption. Hate relevance loss ($\mathcal{L}_{\text{rel}}$) on the other hand, uses only the caption scores directly from the caption scorer. Both use \textit{binary cross-entropy loss} for calculation. While $\mathcal{L}_{\text{cls}}$ predicts whether an image-text pair represents hateful content, $\mathcal{L}_{\text{rel}}$ directly evaluates how `hateful' each caption is and encourages the caption scorer to assign higher scores to hateful captions for hateful images.





These components interact through joint optimization:
\[\mathcal{L}_{\text{total}} = 
\underbrace{\mathcal{L}_{\text{cls}}}_{\text{classification}} + 
\underbrace{\mathcal{L}_{\text{rel}}}_{\text{hate alignment}}\]


The model learns to: 1) weight captions by hate relevance ($\mathcal{S}$), 2) maintain differentiability in selection ($p_i$), and 3) directly connect caption choices to label supervision (\( \mathcal{L}_{\text{rel}} \)), enabling the model to specialize for hate detection while preserving original text understanding through partial parameter updates. These losses are back-propagated to the model for fine-tuning.


\begin{table}[!ht]
    \centering
    \begin{tabular}{c|ccc}
    \toprule
        \textbf{Dataset} & \#\textbf{Training} & \#\textbf{Val} & \#\textbf{Test} \\
        \midrule
        Hateful Memes & $8.5k$ & $500$ & $1k$ \\
        MultiOFF & $445$ & $149$ & $149$ \\
        \bottomrule
    \end{tabular}
    \caption{Dataset Splits. Number of samples allocated to the training, validation (Val), and test sets for the Facebook Hateful Memes (FHM) and MultiOFF datasets used in the experiments.}
    \label{tab:dataset}
\end{table}

\subsection{Experimental Setup} \label{subsec:exp-setup}

We perform experiments in three different settings as described below. Our primary test set is the FHM dataset, which is one of the most challenging meme datasets containing benign confounders, and contains $1k$ test samples. While we acknowledge that this may not be the largest test set, it, however, reflects human-labelled real-world memes from social media platforms. Further, to evaluate the generalizability of our framework, we apply \acro to the problem of offensive language identification (MultiOFF dataset), which has a much smaller test set (Table~\ref{tab:dataset}). Our primary results are over the pre-defined test set of FHM, which also helps us compare our performance with existing methods.

\paragraph{Zero-shot} We perform zero-shot experiments with multiple LVLMs for hate detection, allowing us to report the performance of only the LVLM layer, without any other component of \acro. For these experiments, we provide the original meme image and overlay text as input within the prompt described in the supplementary file.


\paragraph{Projection Layer Fine-tuning} This experiment compares the performance of two VLMs, \cliplarge, and LLM2CLIP-Llama-3.2 (1B-Instruct-CC-Finetuned) with two existing approaches, HateCLIPper-Align~\cite{kumar2022hate}, and RGCL~\cite{mei2023improving} which utilize representation fusion. For fine-tuning both VLMs, we first take the encoder pooler outputs (\textit{i.e.}, the [CLS] token) and map them through linear plus dropout blocks for text and image separately. These mapped representations are then passed through a short stack of ``pre-output'' layers, which apply dropout, linear transformation, and an activation function before the final classification.

\paragraph{\acro} We utilize \internvl, and \gemini to generate candidate captions for \acro. We also perform multiple experiments varying VLM models, and the number of tunable layers ($n$) in the final framework block, while jointly optimizing the caption scorer for hate detection. We perform experiments with \cliplarge, \siglip~\cite{tschannen2025siglip} and the large variant \clipxlmr models \citep{cherti2023reproducible} as they report strong performance on downstream tasks involving vision-language. 


All publicly available models are obtained from the HuggingFace repository, while for \gemini and \gpt, we use their respective APIs. The experiments are conducted using two $24GB$ $A5000$ GPUs. All experiments on both datasets (Table \ref{tab:dataset}) are executed with a batch size of $64$, with a gradient accumulation up to $512$, and a learning rate of $1e-4$ for $30$ epochs with early stopping, with an average runtime of $2$ hours per experiment. 

\paragraph{Evaluation Metrics} To evaluate our models’ ability to flag hateful memes, we report macro‐averaged precision, recall, F1‐score, and accuracy using the Scikit‑learn toolkit~\cite{kramer2016scikit}. We omit AUROC score, since it only measures the ranking of predicted scores and ignores whether those probabilities reflect true hateful content~\cite{lobo2008auc}. Furthermore, it aggregates performance across all possible thresholds, rather than relying on a single decision point used in deployment. It treats false positives and false negatives as equally important, despite missed hateful instances (false negatives) being far more problematic than over‐flagging benign memes~\cite{halligan2015disadvantages}. Consequently, F1 and accuracy offer more meaningful insights into our classifiers’ real‐world effectiveness.


\section{Experimental Results} \label{sec:experiments}

\paragraph{Zero-shot}
From Table \ref{tab:zero-shot}, the zero-shot results show that \gemini \citep{team2023gemini} performs significantly better than the other two LVLMs, achieving an F1 score of $0.756$, while the $8$B variant of the \internvl model \citep{chen2024expanding} achieves $0.679$, and \gpt \citep{hurst2024gpt} achieves $0.703$ F1. To evaluate the viability of LLMs for such a task, we further experimented by providing the generated captions from LVLMs as input to an LLM for zero-shot hate detection. However, the results were significantly inferior to LVLMs in zero-shot.


\begin{table}[!ht]
    \centering
    \resizebox{\columnwidth}{!}{%
    \begin{tabular}{c|cccc}
        \toprule
        \textbf{Model} & \textbf{Acc.} & \textbf{F1} & \textbf{P} & \textbf{R} \\
        \midrule
        \internvl & 0.690 & 0.679 & 0.688 & 0.671 \\
        \gemini & 0.741 & 0.756 & 0.698 & 0.824 \\
        \gpt & 0.692 & 0.703 & 0.666 & 0.745 \\
        \bottomrule
    \end{tabular}
    }%
    \caption{Zero-Shot hate detection performance on the FHM Dataset. Evaluates the performance of various LVLMs (\internvl, \gemini, and \gpt) on the FHM test split without any task-specific fine-tuning. Acc: Accuracy, P: Precision, R: Recall.}
    \label{tab:zero-shot}
\end{table}

\paragraph{Projection Layer Fine-tuning}
As shown in Table \ref{tab:projection-ft}, although the \cliplarge and the newer variant of LLM2CLIP \citep{huang2024llm2clip} models already report good results in other vision–language tasks, simply fine-tuning the projection layers on the generated captions did not suffice. They achieve accuracy scores of $0.720$ and $0.673$, respectively; however, they still fall short compared to existing CLIP-based approaches such as Hate-CLIPper (Acc $0.788$) and RGCL (Acc $0.756$). This suggests that simple projection layer fine-tuning is insufficient for effectively utilizing the caption information. 



\begin{table}[!ht]
    \centering
    \resizebox{\columnwidth}{!}{%
    \begin{tabular}{c|ccccc}
        \toprule
        \textbf{Model} & \textbf{Acc.} & \textbf{F1} & \textbf{P} & \textbf{R} \\
        \midrule
        {\footnotesize \cliplarge} & 0.720 & 0.710 & 0.746 & 0.717 \\[2pt]
        {\footnotesize $\text{LLM2CLIP-Llama-3.2}$} & 0.673 & 0.652 & 0.716 & 0.668 \\[2pt]
        \makecell[c]{\footnotesize Hate-CLIPper - Align} & $0.756^\star$ & -- & -- & -- \\[2pt]
        \makecell[c]{\footnotesize RGCL} & 0.788 & -- & -- & -- \\
        \bottomrule
    \end{tabular}
    }%
    \caption{Projection layer fine-tuning results on the FHM Dataset. Compares performance when fine-tuning only the projection layers of \cliplarge and LLM2CLIP-Llama-3.2-1B. Results are benchmarked against prior methods (Hate-CLIPper-Align, RGCL) using similar lightweight tuning. $^\star$Score reproduced from their code.}
    \label{tab:projection-ft}
\end{table}

\begin{figure}[!ht]
    \centering
    \includegraphics[width=0.95\columnwidth]{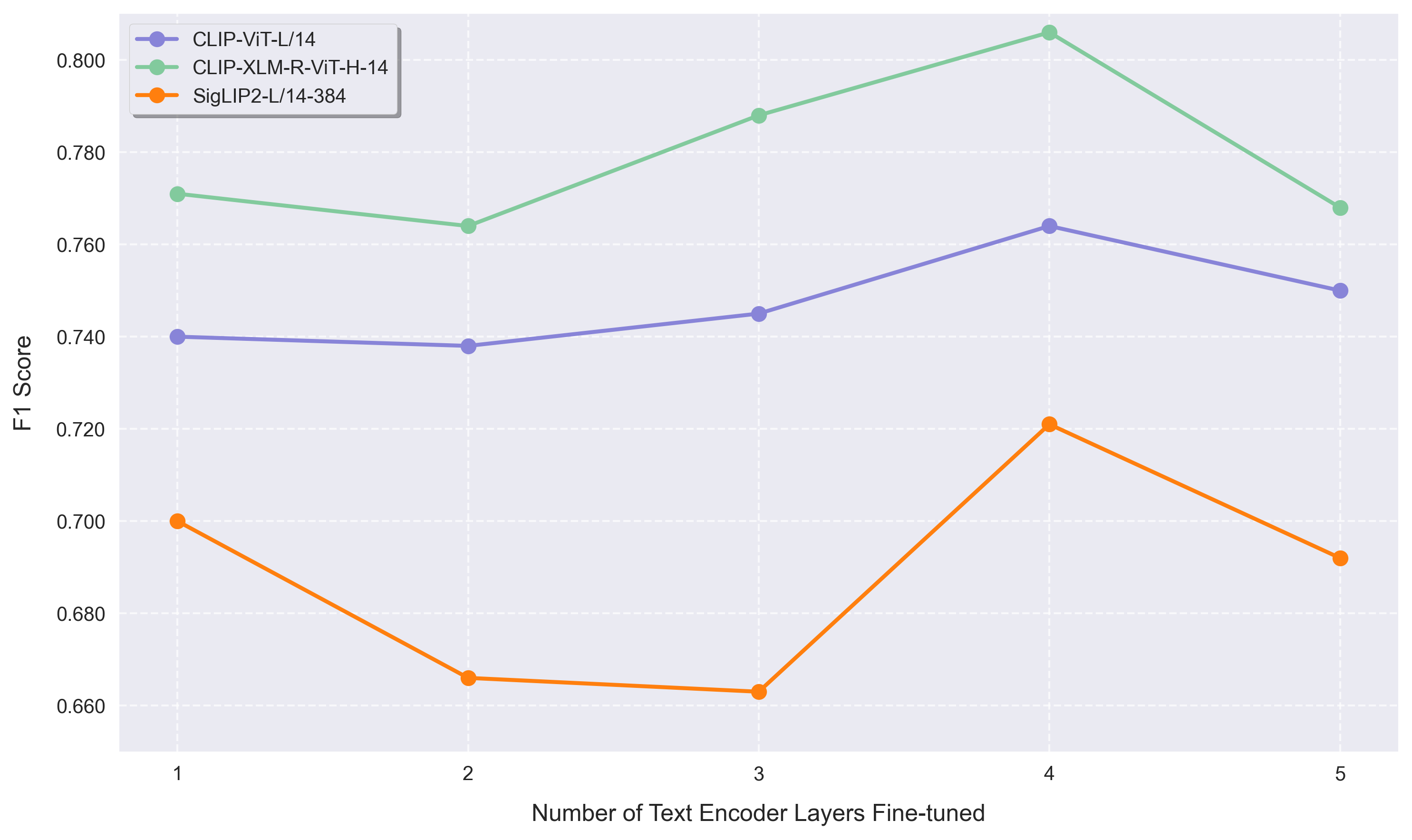}
    \caption{F1 Score vs. the number of fine-tuned text encoder layers ($n$). Illustrates the impact of increasing the number of trainable layers ($n$, from $1$ to $5$) in the text encoder on F1 performance for \cliplarge, \clipxlmr, and \siglip~models within the \acro framework on the FHM dataset. Fine-tuning $4$ layers achieves the best performance for all three models.}
    \label{fig:f1_vs_n}
\end{figure}

\paragraph{\acro's Fine-tuning}
As Table~\ref{tab:encoder-ft} demonstrates, selectively fine-tuning the text encoder yields a clear performance boost compared to full text-encoder fine-tuning in smaller models. In particular, increasing the number of trainable layers to $4$ yields the best F1 score, as visualized in Figure~\ref{fig:f1_vs_n}. This pattern appears for \cliplarge, \siglip, and \clipxlmr, indicating that deeper text encoder updates improve the model’s ability to capture subtle, context-dependent signals in hateful memes for CLIP-based models. The number of trainable layers ($n=4$) may not be universally optimal as layer counts may vary by encoder and task, however, larger fine-tuning is less parameter-efficient. \citet{chen2025rethinking} examine visual layer selection for CLIP models but do not provide a layer-wise analysis for the CLIP text encoder; their findings are empirical and align with our experimental results. With the proposed hate-relevance loss, we observe \acro achieves SoTA performance on the FHM dataset, with an F1 of $0.806$ and an Accuracy of $0.807$.

Pairwise McNemar’s tests~\cite{McNemar_1947} on the FHM test set show that \acro with \clipxlmr~($n=4$) is significantly better than \siglip~($p<0.001$), while its difference from \cliplarge ($n=4$) is not statistically significant ($p=0.8825$); see Appendix D in the supplementary file for details.

Finally, as shown in Table~\ref{tab:multioff-results}, we also evaluated \acro (w/ CLIP-XLM-R-ViT-H-14) with \internvl and \gemini on the MultiOFF dataset to demonstrate the generalizability of our approach. The table also presents the best existing benchmarked results as presented in the original dataset paper for MultiOFF (Baseline). Despite the smaller sample size and different domain focus of MultiOFF, our framework still achieves a notably higher F1 score compared to prior work. This result underlines the broader applicability of our methodology, even when dealing with more limited datasets or varied offensive content scenarios.

\subsection{Discussion} \label{sec:discussion}

Overall, \acro performs at par with the PALI-X-VPD~\cite{hu2023visual} framework, which currently reports SoTA performance over the FHM dataset. It reports an accuracy of $0.808$, while \acro reports an accuracy of $0.807$ and an F1 score of $0.806$. Further analysis using last-layer fine-tuning (Table \ref{tab:encoder-ft}) isolates the impact of the loss components under minimal parameter updates. This suggests that our novel caption scorer and the associated $\mathcal{L}_{\text{rel}}$ provide targeted and effective signals to adapt the model to this specific hate detection task.

\paragraph{Importance of Caption-Scorer and Visual Grounding}   

\begin{figure}[!ht]
    \centering
    \subfloat[\centering True Label: \underline{Hateful}, Prediction: Hateful]{{\includegraphics[width=0.49\columnwidth]{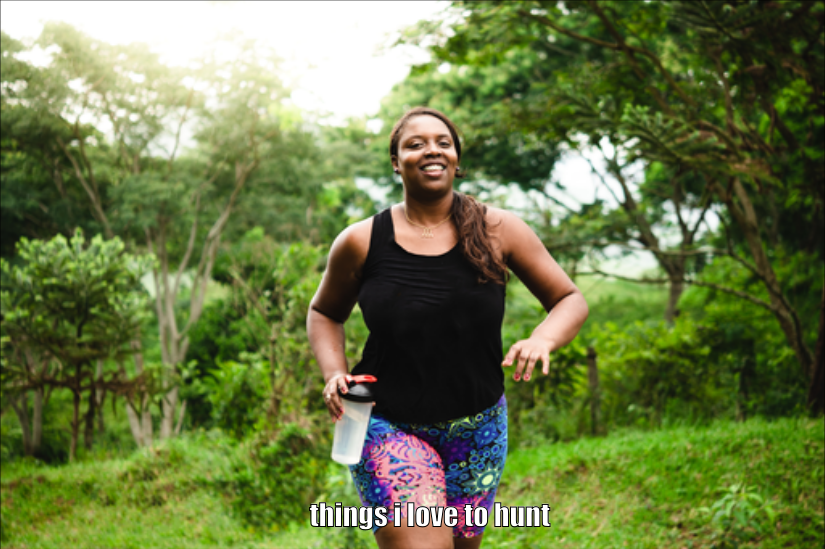} }}%
    \subfloat[\centering True Label: \underline{Not Hateful}, Prediction: Not Hateful]{{\includegraphics[width=0.49\columnwidth]{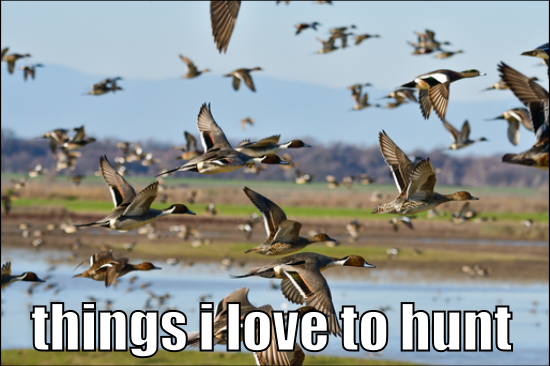} }}%
    \caption{\acro predictions using CLIP-XLM-R-ViT-H-14. (a) A correct classification where subtle hateful objectification is identified. (b) A correctly identified non-hateful meme.}%
    \label{fig:benco_error_analysis_1}%
\end{figure}

Our qualitative analysis highlights the critical roles of visual grounding and caption scoring in \acro. Visual grounding notably reduces hallucinations in smaller models (e.g., \internvl), aiding accurate entity identification. For instance, in meme `07193.png' (Figure~\ref{fig:benco_error_analysis_1}a), visual grounding improved \internvl's basic visual recognition but still missed subtle racial stereotypes implied by the text ``things I love to hunt.'' In contrast, the larger \gemini model generated a caption explicitly recognizing these racial stereotypes, which the caption scorer correctly identified as most hate-relevant (shown in Appendix B of the supplementary file). This targeted caption selection enabled \acro to effectively detect nuanced hate, demonstrating how the combination of robust visual grounding and the caption scorer significantly enhances multimodal hate detection accuracy and interpretability.

\begin{table}[!ht]
    \centering
    \small
    \resizebox{0.95\columnwidth}{!}{%
    \begin{tabular}{c|c|cccc}
        \toprule
        \textbf{Model} & \textbf{FT} & \textbf{Acc.} & \textbf{F1} & \textbf{P} & \textbf{R} \\
        \midrule
        \multirow{2}{*}{\makecell[c]{\footnotesize \textsc{CLIP-ViT} \\ \textsc{L/14}}}
            & \footnotesize T (-1) & 0.743 & 0.740 & 0.757 & 0.745 \\
            & \footnotesize T (-4) & 0.764 & 0.764 & 0.764 & 0.764 \\
        \midrule
        \multirow{2}{*}{\makecell[c]{\footnotesize \textsc{SigLIP2} \\ \textsc{L/16-384}}}
            & \footnotesize T (-1) & 0.705 & 0.700 & 0.730 & 0.708 \\
            & \footnotesize T (-4) & 0.723 & 0.721 & 0.733 & 0.725 \\
        \midrule
        \multirow{2}{*}{\makecell[c]{\footnotesize \textsc{CLIP-XLM-T} \\ \textsc{ViT-H/14}}}
            & \footnotesize T (-1) & 0.774 & 0.771 & 0.794 & 0.776 \\
            & \footnotesize T (-4) & \textbf{0.807} & \textbf{0.806} & \textbf{0.813} & \textbf{0.808} \\
        \bottomrule
    \end{tabular}
    }
    \caption{Performance of \acro fine-tuning on the FHM Dataset with ablations. Evaluates \acro using different VLMs and varying the number of fine-tuned text encoder layers ($n=1$ and $n=4$) using $\mathcal{L}_{\text{cls}} + \mathcal{L}_{\text{rel}}$ losses. The best configuration achieving SoTA-comparable results is marked in bold. FT: Fine-tuning, T: Text, Acc: Accuracy, P: Precision, R: Recall. $(-1)$ and $(-4)$ indicate that last layer and last 4 layers of the text encoder are fine-tuned, respectively.}
    \label{tab:encoder-ft}
\end{table}

\begin{table}[!ht]
    \centering
    \resizebox{0.95\columnwidth}{!}{%
    \begin{tabular}{c|cccc}
        \toprule
        \textbf{Model Configuration} & \textbf{Acc.} & \textbf{F1} & \textbf{P} & \textbf{R} \\
        \midrule
        \makecell[c]{MultiOFF Baseline  \\ \citep{suryawanshi2020multimodal}} & -- & 0.540 & 0.390 & \textbf{0.840} \\[2pt]
        \makecell[c]{\gemini \\ (Zero-shot)} & 0.557 & 0.492 & 0.451 & 0.542 \\[2pt]
        \makecell[c]{\cliplarge \\ (projection FT)} & 0.604 & 0.377 & 0.302 & 0.500 \\[2pt]
        {\acro (ours)} & \textbf{0.678} & \textbf{0.673} & \textbf{0.673} & 0.681 \\
        \bottomrule
    \end{tabular}
    }%
    \caption{Performance comparison on the MultiOFF Dataset. Evaluates the generalizability of \acro by comparing its performance against the original MultiOFF baseline, zero-shot \gemini, and \cliplarge projection layer fine-tuning (FT). \acro results use the configuration \clipxlmr, $n=4$, w/ ($\mathcal{L}_{\text{cls}} + \mathcal{L}_{\text{rel}}$) losses.}
    \label{tab:multioff-results}
\end{table}

\begin{figure}[!ht]
    \centering
    \subfloat[\centering True Label: \underline{Not Hateful}, Prediction: Not Hateful]{{\includegraphics[width=0.49\columnwidth]{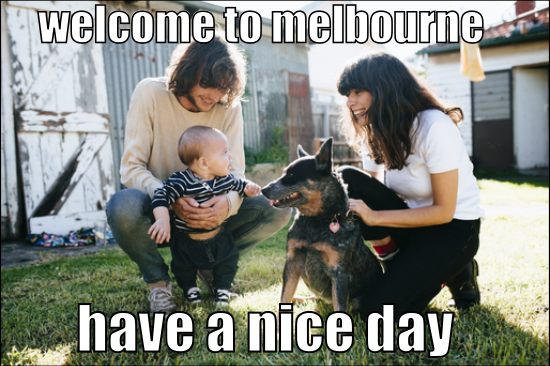} }}%
    \subfloat[\centering True Label: \underline{Hateful}, Prediction: Not Hateful]{{\includegraphics[width=0.49\columnwidth]{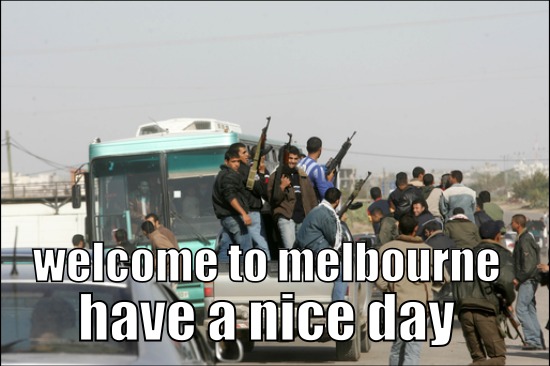} }}%
    \caption{\acro predictions using CLIP-XLM-R-ViT-H-14. (a) A correctly identified non-hateful meme. (b) A misclassification where a hateful meme implying potential violence (guns) is missed, possibly due to visual grounding challenges with small objects.}%
    \label{fig:benco_error_analysis_3}%
\end{figure}

\paragraph{Qualitative Analysis}
We further evaluated the \acro's best output from the \clipxlmr model by conducting qualitative analysis on samples with benign confounders. Figure~\ref{fig:benco_error_analysis_1} shows a confounder pair where the same template can be benign or hateful depending on context. \acro correctly classifies both memes. Visual grounding focuses captioning on the salient entities (\textit{e.g.,} the runner and overlaid text), and the caption scorer prioritizes the more nuanced caption. Figure~\ref{fig:benco_error_analysis_3} contrasts a correct negative with a false negative. In 6a, a common, non-hateful template is correctly rejected. In 6b, the meme implies potential violence, but small, low-salience objects (guns) are not consistently captured by the grounding stage, leading the classifier to mislabel. This points to a remaining limitation: sensitivity to small or partially occluded objects that carry decisive semantic weight. This stems from the current foundation models that could be in the form of hallucinations or failing to find obscured objects in the image (which are difficult to find many a times, even for humans). This limitation may only be addressed as models improve in understanding low-resolution images over time, or when models with more robust patch-level visual understanding catering to the depth within an image are discovered. These examples show that \acro is robust to benign confounders when salient cues are grounded, but can miss cases where key evidence is small or visually subtle.

The supplementary file presents additional examples predicted by \acro, comparing different candidate captions, their scores, and how these eventually influence the classification outcome.

\paragraph{\acro as an Interpretable Framework}
Existing systems, such as HateCLIPper~\cite{kumar2022hate} and RGCL~\cite{mei2023improving}, only produce classification labels, and there is no way for a moderator to understand how the model arrived at that decision. \acro’s modular design makes use of visual grounding, caption generation, and Gumbel-softmax reliant caption scorer to assign probabilities to any candidate captions. This helps the CLIP model pick the best assisting caption to make an informed prediction. Thus, when a moderator looks at the predicted label along with the caption scores and probabilities, they can understand the model’s decision process regardless of a correct or incorrect model prediction.

\paragraph{Computational Efficiency}
To further highlight \acro’s computational efficiency and methodological novelty, we compare it directly against PaLI-X-VPD~\cite{hu2023visual}, the current SoTA approach. As shown in Table~\ref{tab:efficiency-comp}, \acro achieves comparable accuracy to PaLI-X-VPD while requiring drastically fewer computational resources, significantly reducing training time from two days using large TPU clusters to approximately four hours on just two GPUs while incorporating visual grounding models and LVLMs for generating contextually enriched captions. These components not only enhance detection accuracy but also provide interpretability by clearly linking multimodal signals to model predictions. Such efficiency and interpretability advantages position \acro favorably for practical deployment scenarios, especially in resource-constrained environments.

\section{Conclusion and Future Work} \label{sec:conclusion}
In conclusion, we propose the \acro framework that enriches hateful meme detection with visually grounded context augmentation, caption scoring, and parameter-efficient text encoder fine-tuning. Our experiments showed that simply fine-tuning projection layers on strong underlying models does not fully leverage the generated captions. Instead, targeted text encoder tuning substantially improved accuracy and F1 across datasets (FHM and MultiOFF). Notably, our performance is at par with the $55B$ parameter SoTA PALI-X-VPD framework and \acro achieves an accuracy of $0.807$ and an F1 score of $0.806$ while being more efficient. This reflects the framework’s ability to capture subtle linguistic signals, an ability reinforced by visual grounding and enhancing context with LVLMs.


\renewcommand{\arraystretch}{3.0}
\begin{table}[!ht]
    \centering
    \resizebox{0.95\columnwidth}{!}{%
    \begin{tabular}{c|c|c}
        \hline
        \makecell{\textbf{Metric}} & \makecell{\textbf{PaLI-X-VPD} \\ \cite{hu2023visual}} & \makecell{\textbf{\acro{} (ours)}} \\ 
        \hline
        \makecell[c]{Model \\ Size} & \makecell[c]{$55B$ parameters \\ + code gen \\ + visual tools \\ + CoT} & \makecell{$1.1B$ parameters (largest variant) \\ + vg (RAM + GroundingDINO) \\
        + cap gen (\internvl \\ + \gemini)} \\ \hline
        \makecell[c]{Hardware \\ Used} & \makecell{128 TPU-v3 \\ / 128 TPU-v5} & \makecell{2 $\times$ NVIDIA A5000 GPUs \\ (24GB each)} \\ \hline
        \makecell[c]{Training \\ Time} & $\sim$2 days & $\sim$4 hours (2h cap gen + 2h training) \\ \hline
        \makecell[c]{Inference \\ Time (avg)} & \makecell{8.9s (4.7s code gen. \\ + 4.2s exec.)} & \makecell{$\sim$8s (5.5s vg \& cap gen \\ + 2.5s single forward pass)} \\ \hline
        Acc (FHM) & 0.808 & 0.807 \\ \hline
        F1 (FHM) & - & 0.806 \\ \hline
    \end{tabular}
    }
    \caption{Efficiency comparison with PaLI-X-VPD}
    \label{tab:efficiency-comp}
\end{table}

\paragraph{Future Work} Although our text encoder fine-tuning strategy already shows strong performance, it still involves tuning millions of parameters, suggesting that scaling up the training set—\textit{e.g.}, incorporating larger datasets such as MMHS150K or combining multiple (recently released) meme datasets could substantially enhance both generalizability and robustness. In addition, we plan to explore the extraction of intermediate-layer representations from encoders, since different layers may capture distinct linguistic or semantic nuances. Notably, all these directions fit naturally into our proposed framework: by treating it as the core pipeline, such refinements remain modular to apply to any meme dataset.


\section{Limitations and Ethical Considerations}

Our work relies on public datasets and publicly available pre-trained models; therefore, no ethical review was necessary. Nevertheless, these models can inherit biases from their original training data, potentially yielding skewed or harmful judgments for under-represented groups. From an ethical standpoint, responsible usage of this system would require human oversight to prevent over-censorship of benign content and to ensure a review of sensitive or ambiguous cases. Additionally, large encoder fine-tuning can be computationally expensive, limiting real-time or resource-constrained deployments.


\bibliography{aaai2026}

\end{document}


\appendix

\section{Prompt Template}   \label{sec:appendix-prompt}

\subsection{Caption Generation} \label{subsec:appendix-capgen}
\begin{lstlisting}[numbers=none]
grounding_prompt = "Key elements in the image:"
if grounding_info:
    for obj, boxes in grounding_info.items():
        confidence = max(box[4] for box in boxes)
        if confidence > 0.7:
            certainty = "high certainty"
        elif confidence > 0.5:
            certainty = "moderate certainty"
        else:
            certainty = "low certainty"
        grounding_prompt += f" {obj} (detected with {certainty}, count: {len(boxes)}),"
    grounding_prompt = grounding_prompt.rstrip(',') + "."
else:
    grounding_prompt += " No specific object information available."

prompt = f"""<image>
{grounding_prompt}
Task: Analyze this meme image using the above grounding information and generate a **single caption** (under 77 tokens) suitable for CLIP fine-tuning.

The caption should:
- Describe the main visual elements (people, objects, and setting)
- Mention key identifying features (e.g., race, gender, religion, disability, nationality) if relevant
- Summarize the text overlay (if short) or explain its meaning concisely
- Describe any societal implications or stereotypes referenced in the image
- Avoid subjective interpretations or judgmental labels (e.g., 'hateful', 'offensive')

Additionally, if the meme references a specific group (race, religion, sex, etc.), ensure it is explicitly mentioned.
If the meme implies attitudes like exclusion, mocking, contempt, or inferiority, briefly describe how the text and image interact to convey this.

Format the response as:
Caption: [Generated caption here]
"""
\end{lstlisting}

\subsection{Zero-shot Evaluation} \label{subsec:appendix-zshot}
\begin{lstlisting}[numbers=none]
analysis_prompt = f"""Analyze this meme by considering both the visual elements and the text content: "{text}"

Focus on:
1. What is shown in the image
2. How the text relates to the image
3. The overall message or intent
4. Whether this combination promotes bias, discrimination, or harmful stereotypes

Provide your analysis in a brief paragraph."""

response = predict_with_retry(model, analysis_prompt, image)
analysis = response.text

decision_prompt = f"""Based on the following analysis of a meme:
{analysis}

Determine if this meme contains hateful content by considering:
- Does it promote discrimination or prejudice?
- Does it target specific groups based on protected characteristics?
- Does it use harmful stereotypes or derogatory language?
- Is the combination of image and text intended to cause harm or offense?

Respond with ONLY 'hateful' or 'not hateful', followed by a brief explanation after a semicolon."""
\end{lstlisting}

\section{Impact of Visual Grounding} \label{sec:appendix-vg}

For image 07193.png (\textbf{Figure 5a}):

\textit{InternVL w/o visual grounding}: A woman and a girl are jogging in a lush green forest. The woman is wearing a black tank top and colorful leggings, while the girl is in a black top and patterned leggings. They both have water bottles. The text overlay reads ``things I love to hunt'', suggesting a playful or adventurous theme. The image does not reference any specific group or imply exclusion, but the text could be interpreted as a light-hearted expression of enjoyment in outdoor activities.
\\

\textit{InternVL w/ visual grounding}: Woman running outdoors, smiling, holding a bottle, surrounded by greenery. Text overlay reads, ``things i love to hunt''. The image and text promote fitness activities and personal passions without references to any specific group.
\\

As we can see from the above example, visual grounding helps in correctly identifying the entities in the image and reduces hallucination in LVLMs while generating the captions. However, for a much larger model like Gemini-2.5-Flash, this does not make much of a difference and generates a similar caption in both cases, as shown below.
\\

\textit{Gemini Caption}: A Black woman in legging shorts and a tank top runs outdoors with a water bottle. The text ``things I love to hunt'' overlays the image, potentially referencing stereotypes about Black women.
Gemini has a much better understanding of the interaction between the image and text compared with InternVL (even with visual grounding). The caption scorer picks this up and it selects the right caption for meme classification.

\section{Caption Selection Output} \label{sec:appendix-caption}

\textbf{Hateful Memes (FHM) dataset}
\begin{lstlisting}[numbers=none]
Image img/07193.png (Figure 5a):  
Prediction: Hateful (Probability: 0.0026)
True Label: Hateful  

Caption Scores:

1. Text: things i love to hunt

   - Score: -6.7855 | Raw Prob: 0.0228 | Gumbel Prob: 0.000

2. InternVL Caption: Woman running outdoors, smiling, holding a bottle, surrounded by greenery. Text overlay reads, "things i love to hunt." The image and text promote fitness activities and personal passions without references to any specific group.

   - Score: -5.4245 | Raw Prob: 0.0891 | Gumbel Prob: 0.000

3. Gemini Caption: A Black woman in legging shorts and a tank top runs outdoors with a water bottle. The text "things I love to hunt" overlays the image, potentially referencing stereotypes about Black women.

   - Score: -3.1250 | Raw Prob: 0.8881 | Gumbel Prob: 1.000

Selected Caption Index: 3
\end{lstlisting}

\vspace{0.5cm}

\textbf{MultiOFF dataset}
\begin{lstlisting}[numbers=none]
Image HnYtR6j.png:
Label: 1.0

Caption Scores (Hate Relevance):
1. Hate Score: 0.405 | Raw Prob: 0.319 | Gumbel Prob: 0.000
   Caption: POLICE SAYS HE 'S UPSET DONALD TRUMP STATED WE ARE ONLY GETTING THUGS AND CRIMINALS FROM MEXICO . BURNS COP CAR ... YOU KNOW ... TO PROVE TRUMP WRONG mematic.net 
2. Hate Score: 0.488 | Raw Prob: 0.346 | Gumbel Prob: 0.000

   Caption: The image shows a person in a protest holding a flaming object, with police in the background. The text claims the person is upset about Donald Trump's statement on immigration, leading to the burning of a police car to disprove it. The meme uses exaggeration and sarcasm to critique Trump's views.
   
3. Hate Score: 0.455 | Raw Prob: 0.335 | Gumbel Prob: 1.000
   Caption: Meme shows a man holding a stick up with police in riot gear behind him. Text references Donald Trump's comments on Mexicans & burning a police car to disprove him, conveying sarcasm.

Selected Caption (3): Meme shows a man holding a stick up with police in riot gear behind him. Text references Donald Trump's comments on Mexicans & burning a police car to disprove him, conveying sarcasm.

\end{lstlisting}

\section{Statistical Significance Test} \label{sec:appendix-stats}

\paragraph{McNemar's Test} To assess whether the observed performance differences between pairs of models are statistically significant, we employed McNemar's test\footnote{McNemar, Q. (1947) ‘Note on the Sampling Error of the Difference Between Correlated Proportions or Percentages’, Psychometrika, 12(2), pp. 153–157. doi:10.1007/BF02295996.}. This non-parametric test is suitable for comparing the error rates of two classifiers on the same dataset.

The test operates on a $2\times2$ contingency table that tabulates the outcomes where the two models disagree. The cells of the table are:
$n_{10}$: The number of instances correctly classified by Model 1 but incorrectly by Model 2.
$n_{01}$: The number of instances incorrectly classified by Model 1 but correctly by Model 2.
The null hypothesis ($H_0$) of the test is that the two models have the same error rate, which implies that the number of disagreements should be equal (\textit{i.e.,} $n_{10}=n_{01}$). The test calculates a chi-squared ($\chi^2$) statistic based on these disagreement counts. A low p-value (typically $<0.05$) allows us to reject the null hypothesis, suggesting that the difference in performance between the two models is statistically significant.

\begin{table}[!ht]
\centering
\resizebox{0.95\columnwidth}{!}{%
\begin{tabular}{c|cccc}
    \hline
    \textbf{Model Comparison} & \textbf{n\textsubscript{10}} & \textbf{n\textsubscript{01}} & \textbf{Statistic ($\chi^2$)} & \textbf{p-value} \\
    \hline
    \makecell{\clipxlmr \textit{vs.} \\ \cliplarge} & 93 & 90 & 0.02 & 0.8825 \\
    \makecell{\clipxlmr \textit{vs.} \\ \siglip}    & 197 & 104 & 28.12 & $<0.0001$ \\
    \makecell{\cliplarge \textit{vs.} \\ \siglip}    & 172 & 82 & 31.19 & $<0.0001$ \\
    \hline
\end{tabular}
}
\caption{Pairwise McNemar's Test for Statistical Significance. The test compares the error rates between models. A low p-value ($<0.05$) indicates a statistically significant difference in performance.}
\label{tab:mcnemar}
\end{table}

\footnotesize  
\onecolumn

\section{More Benign Confounder Examples} \label{sec:appendix-benco}

\begin{longtable}{p{0.15\linewidth} p{0.8\linewidth}}
\toprule
\textbf{Image} & \textbf{Meme Details}\\
\midrule
\endfirsthead

\toprule
\textbf{Image} & \textbf{Meme Details (continued)}\\
\midrule
\endhead

\bottomrule
\endfoot

\endlastfoot

{\includegraphics[width=\linewidth, valign=t]{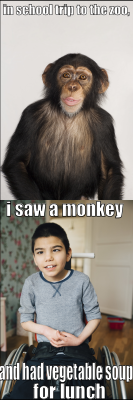}} & 
\textbf{Meme 1: img/63502.png}\newline
\textbf{Prediction}: Hateful (Probability: 0.9984)\newline
\textbf{True Label}: Non-hateful\newline

\textbf{Meme Text:}\newline
Text: in school trip to the zoo, i saw a monkey and had vegetable soup for lunch\newline
Score: -0.8900\newline
Probability: 0.0002\newline

\textbf{InternVL Caption:}\newline
Text: The meme features a child in a wheelchair with a text overlay about a school trip to the zoo. The top image shows a chimpanzee, and the bottom image shows the child. The text humorously suggests that the child saw a monkey and had vegetable soup for lunch, playing on the similarity in appearance between the child and the chimpanzee. This meme uses a visual pun to create a light-hearted joke, but it may be considered offensive to some as it relies on a stereotype of people with disabilities.\newline
Score: 7.7130\newline
Probability: 0.9978\newline
*** Selected as Best Caption ***\newline

\textbf{Gemini Caption:}\newline
Text: A meme showing a monkey and a boy in a wheelchair, with text implying they are related to a zoo trip; The implication mocks the child by equating him to the monkey.\newline
Score: 1.5134\newline
Probability: 0.0020\\
\midrule

\includegraphics[width=\linewidth, valign=t]{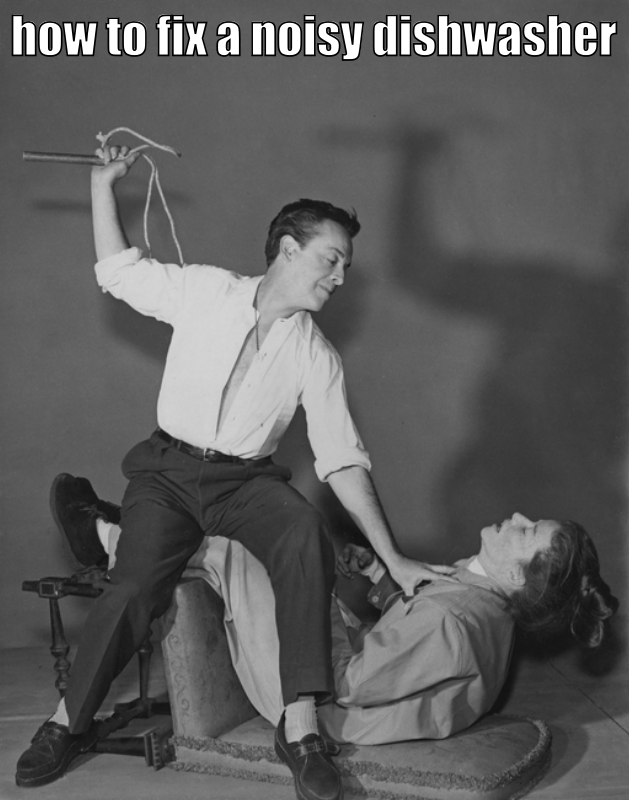} &
\textbf{Meme 2: img/10476.png}\newline
\textbf{Prediction}: Hateful (Probability: 0.9967)\newline
\textbf{True Label}: Hateful\newline

\textbf{Meme Text:}\newline
Text: how to fix a noisy dishwasher\newline
Score: -4.3268\newline
Probability: 0.0000\newline

\textbf{InternVL Caption:}\newline
Text: A man is sitting on a woman's lap, holding a rope, with the text ``how to fix a noisy dishwasher'' overlaid. The image humorously suggests a non-literal solution to a household problem, potentially implying a gender stereotype.\newline
Score: 5.8544\newline
Probability: 0.0175\newline

\textbf{Gemini Caption:}\newline
Text: A black and white image shows a man holding a weapon over a woman lying down. Text reads ``how to fix a noisy dishwasher,'' implying the woman is the dishwasher. This depicts domestic violence towards women in a humorous and belittling way.\newline
Score: 9.8821\newline
Probability: 0.9825\newline
*** Selected as Best Caption ***\\
\midrule

\includegraphics[width=\linewidth, valign=t]{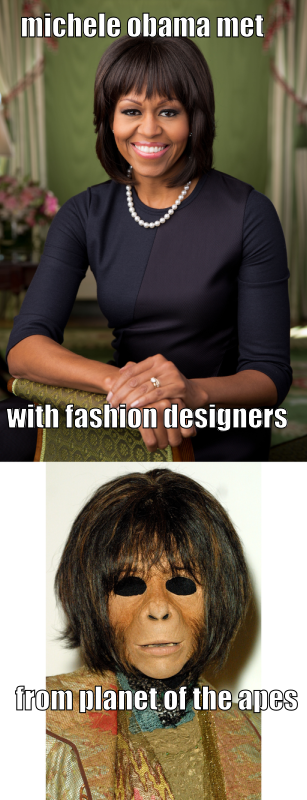} &
\textbf{Meme 3: img/85630.png}\newline
\textbf{Prediction}: Hateful (Probability: 0.9439)\newline
\textbf{True Label}: Hateful\newline

\textbf{Meme Text:}\newline
Text: michele obama met with fashion designers from planet of the apes\newline
Score: 3.6339\newline
Probability: 0.0003\newline

\textbf{InternVL Caption:}\newline
Text: The image features a woman in a dark blue dress with a pearl necklace, sitting with her hands clasped. The text overlay reads, ``michele obama met with fashion designers from planet of the apes.'' The bottom half of the image shows a person with a hair wig resembling an ape, with the text ``from planet of the apes.'' The meme uses a humorous comparison, suggesting that the woman's meeting with fashion designers is as unusual as meeting apes from a fictional planet.\newline
Score: 2.7138\newline
Probability: 0.0001\newline

\textbf{Gemini Caption:}\newline
Text: The two-panel meme shows Michelle Obama, a Black woman, above a character from ``Planet of the Apes.'' The text says ``Michele Obama met with fashion designers from Planet of the Apes,'' implying a comparison that could be interpreted as mocking or belittling.\newline
Score: 11.7253\newline
Probability: 0.9996\newline
*** Selected as Best Caption ***\\
\midrule

\includegraphics[width=\linewidth, valign=t]{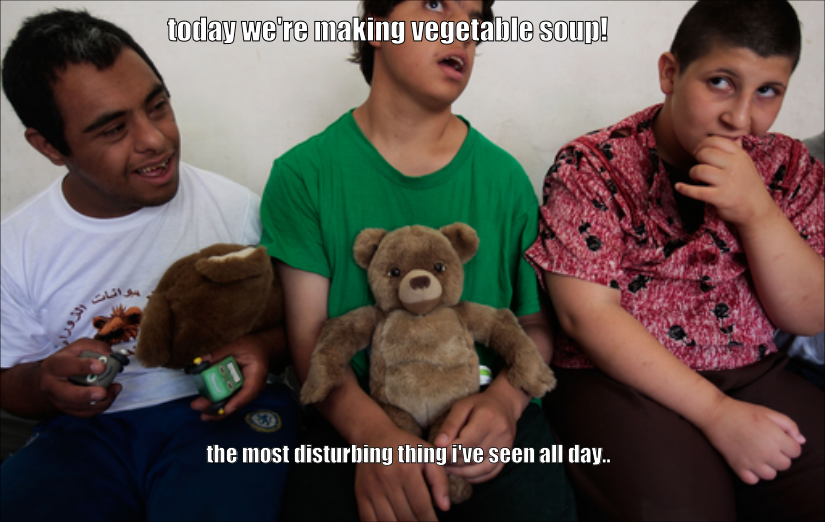} &
\textbf{Meme 4: img/26891.png}\newline
\textbf{Prediction}: Hateful (Probability: 0.8315)\newline
\textbf{True Label}: Hateful\newline

\textbf{Meme Text:}\newline
Text: today we're making vegetable soup! the most disturbing thing i've seen all day.\newline
Score: -5.6092\newline
Probability: 0.0000\newline

\textbf{InternVL Caption:}\newline
Text: Three young boys sit together, one holding a teddy bear, another with a toy, and the third with a bottle. The text overlay reads, ``today we're making vegetable soup!'' and ``the most disturbing thing I've seen all day.'' The image and text juxtapose a seemingly innocent activity with a negative comment, potentially mocking or belittling the boys' interests or the context of the situation.\newline
Score: 1.4312\newline
Probability: 0.0167\newline

\textbf{Gemini Caption:}\newline
Text: A meme shows three individuals who appear to have developmental disabilities, each holding a toy; two hold stuffed toys. The text overlay reads, ``today we're making vegetable soup! the most disturbing thing I've seen all day..'' The meme conveys a sense of mockery or contempt towards individuals with developmental disabilities.\newline
Score: 5.5057\newline
Probability: 0.9833\newline
*** Selected as Best Caption ***\\
\midrule

\includegraphics[width=\linewidth, valign=t]{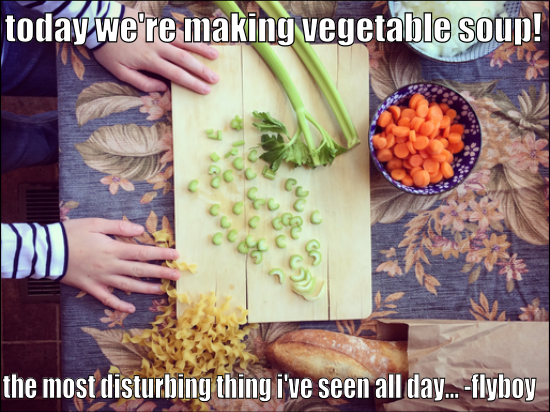} &
\textbf{Meme 5: img/36128.png}\newline
\textbf{Prediction}: Non-hateful (Probability: 0.0015)\newline
\textbf{True Label}: Non-hateful\newline

\textbf{Meme Text:}\newline
Text: today we're making vegetable soup! the most disturbing thing i've seen all day... -flyboy\newline
Score: -4.1946\newline
Probability: 0.1942\newline

\textbf{InternVL Caption:}\newline
Text: In the image, two people are preparing vegetable soup on a cutting board, with carrots, peas, and celery visible. The text overlay reads, ``today we're making vegetable soup!'' and ``the most disturbing thing I've seen all day... -flyboy.'' The meme humorously juxtaposes the wholesome activity of cooking with a sarcastic comment, potentially mocking the idea of flyboy stereotypes.\newline
Score: -4.0403\newline
Probability: 0.2266\newline

\textbf{Gemini Caption:}\newline
Text: A person prepares vegetables, including carrots, celery, and noodles on a cutting board to make vegetable soup. Text reads ``today we're making vegetable soup! the most disturbing thing i've seen all day... -flyboy''. The humor implies the soup ingredients are undesirable to the speaker.\newline
Score: -3.1014\newline
Probability: 0.5793\newline
*** Selected as Best Caption ***\\
\midrule

\end{longtable}